\documentclass[conference]{IEEEtran}
\IEEEoverridecommandlockouts
\usepackage{cite}
\usepackage{amsmath,amssymb,amsfonts}
\usepackage{algorithmic}
\usepackage{graphicx}
\usepackage{textcomp}
\usepackage[table]{xcolor}
\usepackage[most]{tcolorbox} 

\usepackage{hyperref}
\usepackage{booktabs}
\usepackage{multirow}
\usepackage{algorithm}
\usepackage{algorithmic}

\newcommand{\chg}[1]
{%
  {\scriptsize
  \ifdim #1 pt > 0pt
    \textcolor{red}{(#1)}%
  \else
    \textcolor{green!60!black}{(#1)}%
  \fi}
}
\tcbset{
  myboxstyle/.style={
    colback=gray!20,     
    colframe=black!70,    
    coltitle=white,       
    fonttitle=\small\bfseries,  
    fontupper=\footnotesize\itshape,
    boxrule=0.8mm,       
    arc=1mm,               
    boxsep=1mm,             
    left=1mm,              
    right=1mm,              
    top=1mm,               
    bottom=1mm,             
    toptitle=0mm,           
    bottomtitle=0mm,     
    enhanced,                  
  }
}

\def\BibTeX{{\rm B\kern-.05em{\sc i\kern-.025em b}\kern-.08em
    T\kern-.1667em\lower.7ex\hbox{E}\kern-.125emX}}
\begin{document}

\title{Improving Few-Shot Change Detection Visual Question Answering via Decision-Ambiguity-guided Reinforcement Fine-Tuning} 

\author{
\IEEEauthorblockN{
Fuyu Dong\textsuperscript{1},
Ke Li\textsuperscript{1},
Di Wang\textsuperscript{1},
Nan Luo\textsuperscript{1},
Yiming Zhang\textsuperscript{2},
Kaiyu Li\textsuperscript{3},
Jianfei Yang\textsuperscript{1},
Quan Wang\textsuperscript{1}
}
\IEEEauthorblockA{
\textsuperscript{1}School of Computer Science and Technology, Xidian University, Xi'an, China\\
\textsuperscript{2}Department of Mathematics, University of California, San Diego, USA\\
\textsuperscript{3}School of Software Engineering, Xi'an Jiaotong University, Xi'an, China\\
}
}

\maketitle

\begin{abstract}
Change detection visual question answering (CDVQA) requires answering text queries by reasoning about semantic changes in bi-temporal remote sensing images.
A straightforward approach is to boost CDVQA performance with generic vision-language models via supervised fine-tuning (SFT).
Despite recent progress, we observe that a significant portion of failures do not stem from clearly incorrect predictions, but from decision ambiguity, where the model assigns similar confidence to the correct answer and strong distractors. 
To formalize this challenge, we define \emph{Decision-Ambiguous Samples} (DAS) as instances with a small probability margin between the ground-truth answer and the most competitive alternative. 
We argue that explicitly optimizing DAS is crucial for improving the discriminability and robustness of CDVQA models. 
To this end, we propose DARFT, a Decision-Ambiguity-guided Reinforcement Fine-Tuning framework that first mines DAS using an SFT-trained reference policy and then applies group-relative policy optimization on the mined subset. 
By leveraging multi-sample decoding and intra-group relative advantages, DARFT suppresses strong distractors and sharpens decision boundaries without additional supervision. 
Extensive experiments demonstrate consistent gains over SFT baselines, particularly under few-shot settings.
\end{abstract}

\begin{IEEEkeywords}
Change detection visual question answering, remote sensing, vision-language models, reinforcement fine-tuning, decision ambiguity
\end{IEEEkeywords}

\section{Introduction}
\label{sec:intro}
Change Detection Visual Question Answering (CDVQA) is a challenging vision-language reasoning task that requires models to infer semantic changes from bi-temporal remote sensing images.
Specifically, given a pair of images captured at different times with a text query, a CDVQA model must not only determine \emph{whether changes have occurred}, but also infer \emph{``what has changed"}.
This capability is crucial for various real-world applications such as urban development monitoring, disaster assessment, and environmental analysis~\cite{tian2024alignment,wang2024kernel,li2025dynamicearth,liu2025remote}.

Despite its practical importance, CDVQA remains relatively under-explored.
Yuan \emph{et al.}~\cite{yuan2022change} first formulated the CDVQA task, constructed a benchmark dataset, and proposed a Siamese-based architecture to enable cross-modal interaction between language and bi-temporal visual features.
Subsequent work by Li \emph{et al.}~\cite{li2024show} introduced explicit change grounding mechanisms to improve answer reliability and interpretability, together with the QAG-360K dataset, further advancing this research direction.
More recently, generic vision-language models (VLMs) have shown promising capabilities in multi-temporal geospatial reasoning through supervised fine-tuning (SFT) on remote sensing datasets~\cite{xuan2025dynamicvl,liu2024change,li2024unirs}.
These advances naturally raise the following question: \emph{Can generic VLMs, when adapted via standard SFT, solve the CDVQA task reliably and robustly?}

However, our empirical results suggest that the answer is non-trivial.
Even after SFT, introducing generic VLMs yields only limited performance gains on CDVQA, and fails to establish a stable advantage over models specifically designed for this task.
To understand the underlying cause of this phenomenon, we conduct a systematic analysis of model predictions.
Surprisingly, many prediction failures cannot be attributed to obvious reasoning errors or missing key visual evidence.
Instead, errors tend to occur in more subtle cases, where the correct answer and one or more incorrect options receive highly similar prediction scores.
In such cases, the model's decision lies near an unstable boundary, where multiple candidate answers can be supported by visually confusable change cues, rendering the prediction highly sensitive to minor perturbations.

\begin{figure}[!t]
    \centering
    \includegraphics[width=\linewidth]{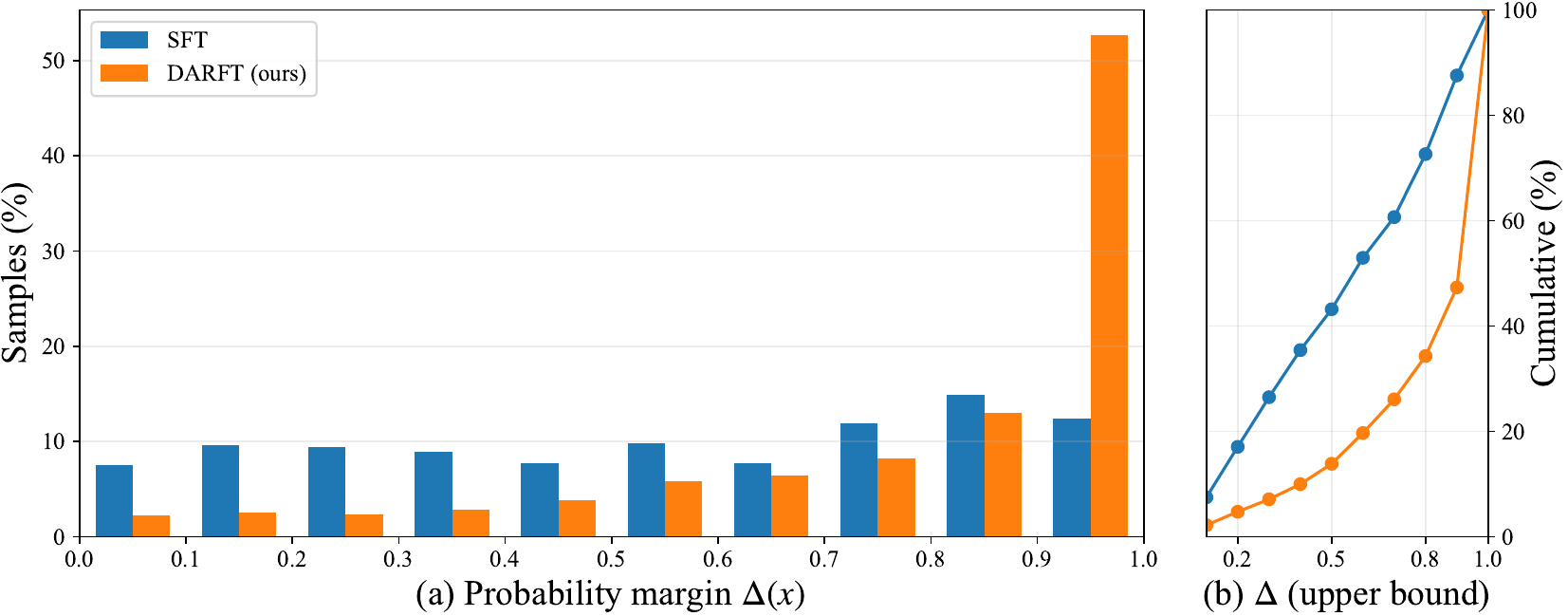}
    \vspace{-6mm}
    \caption{Decision ambiguity distribution before and after DARFT.
    (a) Histogram of decision ambiguity scores $\Delta(x)$, where $\Delta(x)$ denotes the probability margin between the ground-truth answer and the strongest competing distractor.
    (b) Corresponding cumulative distribution functions of $\Delta(x)$.
    Compared to SFT, the proposed DARFT substantially shifts the distribution toward larger margins, indicating reduced decision ambiguity and sharper decision boundaries.}
    \label{fig:teaser}
\end{figure}

To quantitatively characterize this phenomenon, we introduce a sample-level measure of \emph{decision ambiguity}.
Specifically, we first obtain a reference policy $\pi_{\theta_0}$ via supervised fine-tuning and apply it to score each training instance.
For each sample $(x, y)$, the model produces a distribution over candidate answers $p_0(a \mid x) = \pi_{\theta_0}(a \mid x)$.
We define the strongest competing incorrect option as $\hat{a} = \arg\max_{a \in \mathcal{A} \setminus \{y\}} p_0(a \mid x).$
The ambiguity score is then defined as the absolute probability margin between the correct answer and this strongest competitor: $\Delta(x) = \left| p_0(y \mid x) - p_0(\hat{a} \mid x) \right|.$
When $\Delta(x) < \tau$, we regard the sample as a \emph{decision-ambiguous sample} (DAS), indicating that the model's prediction is made under high uncertainty and lies close to the decision boundary.

As shown in Figure.~\ref{fig:teaser}, we visualize this phenomenon by comparing the answer confidence distributions of correctly and incorrectly predicted samples.
The results show that a large portion of prediction errors concentrate in regions with small confidence margin, suggesting a strong association between prediction failures and elevated decision ambiguity.
Importantly, DAS are neither equivalent to misclassified samples nor caused by annotation noise.
Rather, they reflect an intrinsic ambiguity in CDVQA, particularly in fine-grained change detection scenarios where multiple candidate answers remain plausible given visually confusable evidence.

These observations expose a key limitation of existing training paradigms.
Standard supervised fine-tuning (SFT) typically optimizes global likelihood or accuracy, implicitly assuming that each sample admits a single, clearly dominant correct answer~\cite{chu2025sft,ouyang2022training}.
Under this assumption, decision-ambiguous samples are treated indistinguishably from easy ones, and the structured competition between the correct answer and strong distractors is not explicitly modeled.
Existing efforts to improve performance on difficult or uncertain samples generally follow two directions: hard example mining~\cite{smirnov2018hard} and uncertainty-aware modeling~\cite{liang2025uncertainty}.
However, the former mining usually relies on misclassification as the criterion, and therefore fails to capture samples that are predicted correctly yet remain highly ambiguous.
And the latter, on the other hand, focus on quantifying overall predictive uncertainty, without explicitly modeling the relative ordering and margin structure among competing answer candidates.
As a result, models may achieve strong average performance while remaining fragile on decision-ambiguous samples.

To address this issue, we propose a \textbf{D}ecision-\textbf{A}mbiguity-guided \textbf{R}einforcement \textbf{F}ine-\textbf{T}uning framework, called DARFT.
It can improve the model's discriminative capability on boundary-critical cases by explicitly identifying and optimizing the DAS subset.
Specifically, DARFT first aligns the model with the CDVQA task through SFT and automatically identifies decision-ambiguous samples based on prediction margins.
It then applies group-relative policy optimization (GRPO) on the selected subset, sampling multiple candidate answers under the same visual-question context and constructing relative advantages within each group.
This procedure explicitly reinforces a stable preference for the correct answer while suppressing strong distractors, thereby sharpening the model's decision boundary.
Notably, DARFT requires neither additional human annotations nor external reward models.
Extensive experiments on multiple CDVQA benchmarks, including QAG-360K, demonstrate that DARFT consistently outperforms standard SFT baselines, particularly in few-shot settings.
Our contributions are summarized as follows:
\begin{itemize}
    \item We identify decision ambiguity as a key challenge in CDVQA and introduce the concept of Decision-Ambiguous Samples (DAS) to characterize boundary-critical cases.
    \item We propose DARFT, a decision-ambiguity-guided reinforcement fine-tuning framework that explicitly optimizes DAS to enhance the model's discriminative capability.
    \item We conduct extensive experiments demonstrating that DARFT consistently improves CDVQA performance across multiple few-shot settings, establishing new state-of-the-art results.
\end{itemize}

\section{Methodology}
\label{sec:method}
We propose a \emph{Decision-Ambiguity-Guided Reinforcement Fine-Tuning} framework, called DARFT, for CDVQA.
The core idea of DARFT is to explicitly identify and refine predictions near decision boundaries, where multiple answer options receive comparable confidence.
The framework follows a two-stage training paradigm.
First, SFT aligns a general-purpose vision--language model with the CDVQA task distribution.
Second, the model is further optimized on a subset of decision-ambiguous samples using GRPO, with the goal of stabilizing and sharpening predictions in ambiguous regions of the output space.
Between the two stages, we perform an intermediate ambiguity mining step to construct $\mathcal{D}_{amb}$ using the SFT reference policy $\pi_{\theta_0}$, which does not update model parameters.

\subsection{Problem Formulation}
CDVQA is formulated as a multi-choice reasoning problem over bi-temporal remote sensing images.
Each input instance is represented as $x = (I^{t_1}, I^{t_2}, q),$
where $I^{t_1}$ and $I^{t_2}$ denote images captured at two different time, and $q$ is a natural-language query.
The model is required to select an answer from a fixed option set $\mathcal{A}$, with ground-truth label $y \in \mathcal{A}$.
We model prediction as a conditional policy $\pi_\theta(a \mid x)$ parameterized by $\theta$, and focus on improving its discriminability in regions where competing answer options exhibit similar posterior probabilities.

\begin{figure*}[!t]
    \centering
    \includegraphics[width=\linewidth]{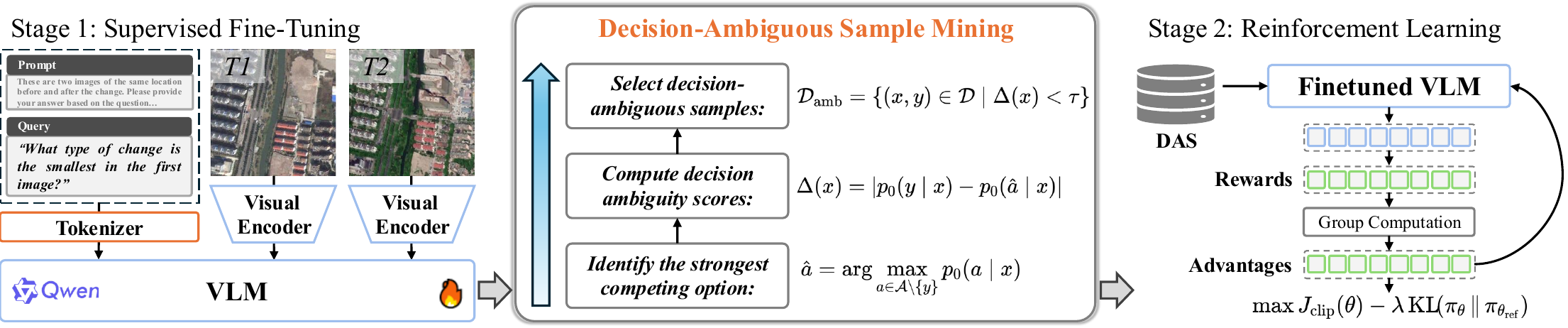}
    \vspace{-6mm}
    \caption{The visualization of the training pipeline of our DARFT framework.}
    \label{fig:framework}
\end{figure*}

\subsection{Model Interface and Output Space}
\noindent \textbf{Prompting.}
To ensure consistent semantic grounding across all answer options, we adopt a unified system prompt that explicitly specifies the meaning of each option.
Following the original dataset protocol, we retain the complete option space of 23 candidates without restricting answer availability.
\begin{tcolorbox}[
    myboxstyle, 
    title=Prompt Template of CDVQA 
]
These are two images of the same location before and after the change. Please provide your answer based on the question. Answer the question with one capital letter, like A, B or C, without anything else. The correspondence between the options and the capital letters is as follows: 0: A, trees: B, bare land: C, low vegetation: D, 20 to 30: E, 70 to 80: F, buildings: G, 0 to 10: H, 10 to 20: I, no: J, yes: K, 40 to 50: L, 50 to 60: M, 30 to 40: N, 90 to 100: O, 60 to 70: P, water: Q, 80 to 90: R, playgrounds: S, road: T, others: U, green house: V, bridge: W. A: 0 is the selection of change ratio, means 0\%.
\end{tcolorbox}

\noindent \textbf{Output and option probabilities.}
Although CDVQA is formulated as a multi-choice task over a discrete option set $\mathcal{A}$, the underlying vision--language model predicts tokens from the full vocabulary $\mathcal{V}$.
Let $z_v(x)$ denote the logit of token $v \in \mathcal{V}$ at the answer position, where $x=(I^{t_1}, I^{t_2}, q)$.
The model defines a token-level distribution via
\begin{equation}
p_\theta(v \mid x) =
\frac{\exp(z_v(x))}{\sum_{v' \in \mathcal{V}} \exp(z_{v'}(x))}.
\end{equation}
Each answer option $a \in \mathcal{A}$ corresponds to a dedicated option token $v(a) \in \mathcal{V}$ (i.e., the letter token).
We therefore define the option probability as
\begin{equation}
p_\theta(a \mid x) \triangleq p_\theta(v(a) \mid x), \quad a \in \mathcal{A}.
\end{equation}
This option probability is consistently used for supervised fine-tuning, decision ambiguity estimation, and reinforcement optimization.

\subsection{Stage I: Task Alignment via Supervised Fine-Tuning}
We first align the vision-language model with the CDVQA task distribution through SFT.
Given the training set $\mathcal{D} = \{(x_i, y_i)\}_{i=1}^{N}$, we optimize the standard cross-entropy objective:
\begin{equation}
\mathcal{L}_{\mathrm{SFT}} 
= - \mathbb{E}_{(x,y)\sim \mathcal{D}} \log p_\theta(y \mid x).
\end{equation}
The resulting policy $\pi_{\theta_0}$ serves as a stable, task-aligned reference.
Beyond providing a strong initialization, it plays a dual role in subsequent stages:
(1) estimating sample-level decision ambiguity, and
(2) acting as a reference policy for KL regularization during reinforcement optimization to prevent excessive policy drift.

\subsection{Decision-Ambiguous Sample Mining}
A key observation in CDVQA is that prediction errors often arise when multiple answer options receive comparable confidence under the same input, indicating weak option separability.
To explicitly identify such cases, we score each training sample using the reference policy $\pi_{\theta_0}$.

For a given $(x,y)$, we compute $p_0(a \mid x) = \pi_{\theta_0}(a \mid x)$ and identify the strongest competing option:
\begin{equation}
\hat{a} = \arg\max_{a \in \mathcal{A} \setminus \{y\}} p_0(a \mid x).
\end{equation}
We then define the decision ambiguity score as the probability margin between the correct answer and the strongest distractor:
\begin{equation}
\Delta(x) = \left| p_0(y \mid x) - p_0(\hat{a} \mid x) \right|.
\end{equation}
Samples with $\Delta(x) < \tau$ are regarded as \emph{decision-ambiguous samples} (DAS) and collected into a subset, denoted as:
\begin{equation}
\mathcal{D}_{\mathrm{amb}} =
\{(x,y) \in \mathcal{D} \mid \Delta(x) < \tau\}.
\end{equation}
Unless otherwise specified, we set $\tau = 0.2$.
This criterion isolates samples whose predictions lie close to the decision boundary and are therefore most sensitive to small perturbations.

\subsection{Stage II: Decision-Ambiguity-Guided GRPO}
We further optimize the model on $\mathcal{D}_{\mathrm{amb}}$ using group-relative policy optimization (GRPO).
Rather than treating each prediction independently, GRPO operates on groups of sampled answers under the same input, enabling explicit comparison among competing options.

\noindent \textbf{Group sampling.}
For each ambiguous input $x$, we sample $K$ candidate answers from the current policy $a^{(k)} \sim \pi_\theta(\cdot \mid x), \quad k=\{1,\dots,K\}$, where $\pi_{\theta_{\mathrm{old}}}$ denotes the old policy used to collect rollouts.
We use a fixed decoding configuration throughout GRPO training, \textit{i.e.}, temperature $T=1.0$ and top-$p=0.9$.

\noindent \textbf{Reward signal.}
Given the discrete nature of CDVQA annotations, we adopt a binary reward reflecting answer correctness which is defined as:
\begin{equation}
r^{(k)} =
\begin{cases}
1, & a^{(k)} = y, \\
0, & \text{otherwise}.
\end{cases}
\end{equation}

\noindent \textbf{Relative advantage construction.}
To emphasize relative performance within each group, rewards are normalized to compute group-relative advantages:
\begin{equation}
A^{(k)} = \frac{r^{(k)} - \mu_r}{\sigma_r + \epsilon},
\end{equation}
where $\mu_r$ and $\sigma_r$ denote the mean and standard deviation of rewards within the group.
This normalization reduces variance across samples and highlights informative gradients when correct answers emerge among strong distractors.

\noindent \textbf{Policy Objective.}
We optimize the policy with a clipped surrogate objective and KL regularization to a reference policy.
Let $\pi_{\theta_{\mathrm{old}}}$ denote the policy before the update and define the probability ratio
\begin{equation}
\rho^{(k)} =
\frac{\pi_\theta(a^{(k)} \mid x)}{\pi_{\theta_{\mathrm{old}}}(a^{(k)} \mid x)}.
\end{equation}
Following GRPO, we maximize the clipped surrogate objective:
\begin{equation}
J_{\mathrm{clip}}(\theta)=
\mathbb{E}\!\left[
\min\!\left(
\rho^{(k)} A^{(k)},
\mathrm{clip}(\rho^{(k)}, 1-\epsilon, 1+\epsilon) A^{(k)}
\right)
\right].
\end{equation}
To limit deviation from the SFT-aligned reference policy $\pi_{\theta_{\mathrm{ref}}}$, we add a KL penalty and maximize:
\begin{equation}
J(\theta)=
J_{\mathrm{clip}}(\theta)
-\lambda\,\mathrm{KL}\!\left(\pi_\theta\,\|\,\pi_{\theta_{\mathrm{ref}}}\right).
\end{equation}
This objective refines decision boundaries on ambiguous samples while constraining policy drift.

\subsection{Inference}
At inference, we use deterministic greedy decoding for fair comparison with prior methods, which is defined as:
\begin{equation}
\hat{y} = \arg\max_{a \in \mathcal{A}} \pi_\theta(a \mid x).
\end{equation}
To analyze robustness to sampling stochasticity, we additionally consider a multi-sample setting.
Given an input $x$, we draw $N{=}20$ \textit{i.i.d.} samples $a_i \sim \pi_\theta(\cdot \mid x)$ using the same sampling hyperparameters as in training.

Under this multi-sample setting, we report two aggregation schemes.
First, \emph{majority voting} predicts the label with the highest empirical frequency, which can be formulated as:
\begin{equation}
\hat{y}_{\mathrm{mv}}(x)
=
\arg\max_{a \in \mathcal{A}}
\sum_{i=1}^{N} \mathbb{I}\!\left[a_i = a\right].
\end{equation}
We report the corresponding \emph{majority-vote accuracy}, which can be represented as:
\begin{equation}
\mathrm{Acc}_{\mathrm{mv}}
=
\frac{1}{|\mathcal{D}|}
\sum_{(x,y)\in \mathcal{D}}
\mathbb{I}\!\left[\hat{y}_{\mathrm{mv}}(x)=y\right].
\label{eq:acc_mv}
\end{equation}
Second, we compute the \emph{mean accuracy} by averaging correctness over the $N$ sampled outputs, which is denoted as:
\begin{equation}
\mathrm{Acc}_{\mathrm{mean}}
=
\frac{1}{|\mathcal{D}|}
\sum_{(x,y)\in \mathcal{D}}
\frac{1}{N}\sum_{i=1}^{N}
\mathbb{I}\!\left[a_i(x)=y\right],
\label{eq:acc_mean}
\end{equation}
where $\mathcal{D}$ denotes the evaluation set and $\mathbb{I}[\cdot]$ is the indicator function.
The DARFT training procedure is summarized in Algorithm~\ref{alg:darft}.
\begin{algorithm}[!t]
\caption{DARFT: Decision-Ambiguity-guided Reinforcement Fine-Tuning for CDVQA}
\label{alg:darft}
\begin{algorithmic}[1]
\REQUIRE Training set $\mathcal{D}=\{(x_i,y_i)\}_{i=1}^N$, option set $\mathcal{A}$, threshold $\tau$, group size $K$,
clip range $\epsilon$, KL weight $\lambda$, 
\ENSURE Optimized policy $\pi_\theta$
\STATE \textbf{Stage I: Supervised Fine-Tuning (SFT)}
\STATE Initialize $\theta \leftarrow \theta_{\text{base}}$
\STATE Minimize $\mathcal{L}_{\mathrm{SFT}}=-\mathbb{E}_{(x,y)\sim\mathcal{D}}\log p_\theta(y\mid x)$
\STATE Set reference policy $\pi_{\theta_0}\leftarrow \pi_{\theta}$
\STATE \textbf{Decision-Ambiguous Sample Mining}
\STATE $\mathcal{D}_{\mathrm{amb}} \leftarrow \emptyset$
\FOR{each $(x,y)\in\mathcal{D}$}
    \STATE Compute $p_0(a\mid x)$ under $\pi_{\theta_0}$ for all $a\in\mathcal{A}$
    \STATE $\hat{a}\leftarrow \arg\max_{a\in\mathcal{A}\setminus\{y\}} p_0(a\mid x)$
    \STATE $\Delta(x)\leftarrow \left|p_0(y\mid x)-p_0(\hat{a}\mid x)\right|$
    \IF{$\Delta(x)<\tau$}
        \STATE $\mathcal{D}_{\mathrm{amb}} \leftarrow \mathcal{D}_{\mathrm{amb}} \cup \{(x,y)\}$
    \ENDIF
\ENDFOR
\STATE {\textbf{Stage II: GRPO with PPO-style clipping on $\mathcal{D}_{\mathrm{amb}}$}}
\WHILE{not converged}
    \STATE $\theta_{\mathrm{old}} \leftarrow \theta$
    \STATE Rollout buffer $\mathcal{B}\leftarrow \emptyset$
    \FOR{each $(x,y)\in\mathcal{D}_{\mathrm{amb}}$ }
        \STATE Sample $\{a^{(k)}\}_{k=1}^{K}$ with $a^{(k)}\sim \pi_{\theta_{\mathrm{old}}}(\cdot\mid x)$
        \FOR{$k=1$ \TO $K$}
            \STATE $r^{(k)} \leftarrow \mathbb{I}[a^{(k)}=y]$
        \ENDFOR
        \STATE $\mu_r \leftarrow \frac{1}{K}\sum_{k=1}^{K} r^{(k)}$
        \STATE $\sigma_r \leftarrow \sqrt{\frac{1}{K}\sum_{k=1}^{K}\left(r^{(k)}-\mu_r\right)^2}$

        \FOR{$k=1$ \TO $K$}
            \STATE $A^{(k)} \leftarrow \frac{r^{(k)}-\mu_r}{\sigma_r+\varepsilon}$
            \STATE Store $(x,a^{(k)},A^{(k)})$ into $\mathcal{B}$
        \ENDFOR
    \ENDFOR
    \FOR{$u=1$ \TO $U$}
        \STATE $J_{\mathrm{clip}} \leftarrow 0$
        \FOR{each $(x,a,A)\in\mathcal{B}$}
            \STATE $\rho \leftarrow \frac{\pi_\theta(a\mid x)}{\pi_{\theta_{\mathrm{old}}}(a\mid x)}$
            \STATE $J_{\mathrm{clip}} \leftarrow J_{\mathrm{clip}} +
            \min\!\Big(\rho A,\ \mathrm{clip}(\rho,1-\epsilon,1+\epsilon)A\Big)$
        \ENDFOR
        \STATE $J_{\mathrm{clip}} \leftarrow \frac{1}{|\mathcal{B}|}J_{\mathrm{clip}}$
        \STATE $J \leftarrow J_{\mathrm{clip}} - \lambda\,\mathrm{KL}(\pi_\theta\ \|\ \pi_{\theta_0})$
        \STATE Update $\theta$ by ascending $\nabla_\theta J$
    \ENDFOR
\ENDWHILE
\STATE \textbf{return} $\pi_\theta$
\end{algorithmic}
\end{algorithm}

\begin{table*}[!t]
\centering
\caption{Comparisons on the QAG-360K test set using deterministic greedy decoding.}
\renewcommand\arraystretch{1.15}
\setlength{\tabcolsep}{1.5pt}
\resizebox{\linewidth}{!}{
\begin{tabular}{l|cccccccc|cc}
\toprule
\textbf{Method} 
& CN & CtW & CfW & IN & DN & LC & SC & CR
& \textbf{AA} & \textbf{OA} \\
\midrule
\multicolumn{3}{l}{\textbf{\textsl{Specialist Methods}}}\\
\midrule
CDVQA~\cite{yuan2022change} 
& 82.25 & 57.81 & 60.44 & 75.41 & 76.76 & 47.67 & 29.27 & 65.20 & 61.85 & 67.91 \\
VisTA~\cite{li2024show} 
& 85.85 & 63.20 & 66.41 & 84.65 & 86.14 & 62.74 & 38.21 & 71.13 & 69.79 & 74.59 \\
\midrule
\multicolumn{2}{l}{\textbf{\textsl{Zero-shot Baselines}}}\\
\midrule
Qwen2.5-VL-3B~\cite{bai2025qwen2}
& 51.37 & 9.93 & 4.69 & 48.22 & 47.55 & 2.74 & 14.96 & 5.14 & 23.08 & 27.44 \\
Qwen2.5-VL-7B~\cite{bai2025qwen2}
& 47.37 & 43.42 & 12.20 & 47.62 & 53.66 & 37.68 & 13.24 & 15.75 & 33.87 & 33.10 \\
\midrule
\multicolumn{3}{l}{\textbf{\textsl{Few-shot Fine-tune (0.5K samples)}}}\\
\midrule
VisTA 
& 68.55 & 42.51 & 36.94 & 66.72 & 64.44 & 28.73 & 19.13 & 47.32 & 41.59 & 52.90 \\
Qwen2.5-VL-3B-SFT
& 77.74 & 53.04 & 52.58 & 71.25 & 57.95 & 35.20 & \textbf{19.57} & 36.50 & 50.48 & 54.52\\
\rowcolor{gray!12}
Qwen2.5-VL-3B-DARFT
& \textbf{78.13} \chg{+0.39} & \textbf{53.99} \chg{+0.95} & \textbf{56.11} \chg{+3.53} & \textbf{74.48} \chg{+3.23} & \textbf{62.45} \chg{+4.50} & \textbf{38.67} \chg{+3.47} & 16.61 \chg{-2.96} & \textbf{54.88} \chg{+18.38} & \textbf{54.42} \chg{+3.94} & \textbf{60.57} \chg{+6.05}\\
\midrule
\multicolumn{3}{l}{\textbf{\textsl{Few-shot Fine-tune (2K samples)}}}\\
\midrule
VisTA 
& 68.29 & 39.26 & 48.34 & 62.14 & 69.25 & 33.06 & 21.90 & 52.23 & 43.83 & 55.13 \\
Qwen2.5-VL-3B-SFT
& \textbf{76.13} & \textbf{53.14} & \textbf{55.23} & \textbf{77.07} & 73.40 & 40.35 & 15.80 & 40.37 & 53.93 & 57.35\\
\rowcolor{gray!12}
Qwen2.5-VL-3B-DARFT
& 75.96 \chg{-0.17} & 46.45 \chg{-6.69} & 54.31 \chg{-0.92} & 76.97 \chg{-0.10} & \textbf{76.70} \chg{+3.30} & \textbf{40.54} \chg{+0.19} & \textbf{18.24} \chg{+2.44} & \textbf{55.88} \chg{+15.51} & \textbf{55.63} \chg{+1.70} & \textbf{61.48} \chg{+4.13} \\
\bottomrule
\end{tabular}}
\label{tab:main}
\end{table*}

\begin{table*}[!t]
\centering
\caption{Comparisons on the QAG-360K test set under \textbf{Mean@20} with multi-sample decoding.}
\renewcommand\arraystretch{1.15}
\setlength{\tabcolsep}{1.5pt}
\resizebox{\linewidth}{!}{
\begin{tabular}{l|cccccccc|cc}
\toprule
\textbf{Method} 
& CN & CtW & CfW & IN & DN & LC & SC & CR
& \textbf{AA} & \textbf{OA} \\
\midrule
\multicolumn{11}{l}{\textbf{Few-shot Fine-tune (0.5K samples)}} \\
\midrule
Qwen2.5-VL-3B-SFT
& 71.65 & 41.67 & 35.38 & 67.59 & 53.44 & 25.24 & 19.22 & 25.82
& 42.50 & 46.74 \\
\rowcolor{gray!12}
Qwen2.5-VL-3B-DARFT
& \textbf{74.65} \chg{+3.00}
& \textbf{49.45} \chg{+7.78}
& \textbf{49.79} \chg{+14.41}
& \textbf{71.62} \chg{+4.03}
& \textbf{60.49} \chg{+7.05}
& \textbf{33.53} \chg{+8.29}
& 18.69 \chg{-0.53}
& \textbf{51.03} \chg{+25.21}
& \textbf{51.16} \chg{+8.66}
& \textbf{57.22} \chg{+10.48} \\
\midrule
\multicolumn{11}{l}{\textbf{Few-shot Fine-tune (2K samples)}} \\
\midrule
Qwen2.5-VL-3B-SFT
& 73.53 & 41.60 & 41.23 & 70.03 & 64.32 & 29.67 & 17.30 & 35.75
& 46.68 & 51.71 \\
\rowcolor{gray!12}
Qwen2.5-VL-3B-DARFT
& \textbf{75.87} \chg{+2.34}
& \textbf{45.73} \chg{+4.13}
& \textbf{53.51} \chg{+12.28}
& \textbf{76.62} \chg{+6.59}
& \textbf{76.28} \chg{+11.96}
& \textbf{40.01} \chg{+10.34}
& \textbf{18.42} \chg{+1.12}
& \textbf{52.09} \chg{+16.34}
& \textbf{54.82} \chg{+8.14}
& \textbf{60.26} \chg{+8.55} \\
\bottomrule
\end{tabular}}
\label{tab:mean20_all}
\end{table*}

\section{Experiments}
\label{sec:exp}
\subsection{Experimental Setup}
\noindent \textbf{Datasets.}
We evaluate DARFT on QAG-360K~\cite{li2024show}, currently the largest and most diverse benchmark for CDVQA.
It covers 8 question categories of varying difficulty, including \emph{Change or Not} (CN), \emph{Change to What} (CtW), \emph{Change from What} (CfW), \emph{Increase or Not} (IN), \emph{Decrease or Not} (DN), \emph{Largest Change} (LC), \emph{Smallest Change} (SC), and \emph{Change Ratio} (CR).
To assess performance in low-data regimes, we construct two few-shot training subsets by randomly sampling 0.5K and 2K instances from the original training split.

\noindent \textbf{Evaluation Metrics.}
To comprehensively evaluate the proposed DARFT framework, we adopt both answer accuracy and spatial grounding metrics.
Following common VQA practice, we report Average Accuracy (AA) and Overall Accuracy (OA) to measure the correctness of predicted answers at the category level and the dataset level, respectively.
To further probe robustness to decoding stochasticity, we additionally report two multi-sample metrics with $N{=}20$ samples per instance: MV@20 (majority-vote accuracy; Eq.~\ref{eq:acc_mv}) and Mean@20 (mean accuracy; Eq.~\ref{eq:acc_mean}), using the same sampling configuration as training (temperature $T{=}1.0$, top-$p{=}0.9$).
For clarity, we highlight improvements over the corresponding SFT baseline in red.

\noindent \textbf{Implementation Details.}
We implement our DARFT framework based on the Qwen2.5-VL-3B model~\cite{bai2025qwen2}, and adopt LoRA~\cite{hu2022lora} for parameter-efficient fine-tuning in both stages. Training is performed on the few-shot training set described in the previous section. For SFT, we use the AdamW~\cite{loshchilov2017decoupled} optimizer for 3 epochs with an effective batch size of 64 and a peak learning rate of $1\times10^{-4}$, where the learning rate follows a cosine decay schedule. We then conduct the GRPO~\cite{shao2024deepseekmath} stage with group size $K=8$ and clip range $\epsilon=0.35$, together with a DAPO-style~\cite{yu2025dapo} dynamic sampling strategy to improve training stability and sample efficiency. The policy is optimized with a peak learning rate of $1\times10^{-5}$ for up to 4 epochs, also using cosine decay, with early stopping once convergence is observed. All experiments are conducted on 4 NVIDIA RTX 4090 GPUs.

\subsection{Main Results}
Table~\ref{tab:main} reports results on QAG-360K under deterministic greedy decoding.
We compare DARFT with specialist CDVQA models, zero-shot VLM baselines, and SFT of Qwen2.5-VL under different data regimes.
Zero-shot VLMs perform poorly on fine-grained change categories, particularly \emph{Change to What}, \emph{Change from What}, and \emph{Change Ratio}, suggesting that such queries require task-specific adaptation beyond generic vision--language pretraining.
While SFT substantially improves overall accuracy, noticeable gaps remain on ambiguity-prone categories, where semantically close options are often supported by visually confusable evidence.

By contrast, DARFT consistently improves upon SFT across both 0.5K and 2K few-shot settings, with the largest gains on fine-grained categories that exhibit high decision ambiguity.
Notably, \emph{Change Ratio} shows the most substantial improvement: \textbf{+18.38} in the 0.5K setting and \textbf{+15.51} in the 2K setting.
These results support our hypothesis that explicitly optimizing decision-ambiguous samples sharpens the model's preference for correct answers over strong distractors.
We also observe minor drops on a small number of categories in certain regimes (\emph{e.g.,} \emph{Change to What} under 2K), which is expected since ambiguity-guided optimization redistributes learning emphasis toward resolving highly confusable instances; category-wise fluctuations can further arise from sampling variance and differences in ambiguity composition across subsets.
Nevertheless, DARFT consistently improves AA and OA, and its largest gains concentrate on fine-grained categories, indicating improved robustness in the most critical region, near decision boundaries where standard SFT remains fragile.

\subsection{Decision Stability under Multi-sample Inference.}
To further examine the robustness of model decisions, we evaluate multi-sample inference settings and report Mean@20 and MV@20 results.
As shown in Table~\ref{tab:mean20_all}, DARFT consistently outperforms SFT under Mean@20 across all categories and data regimes.
The improvements are especially significant for fine-grained and ambiguity-prone categories, with \emph{Change Ratio} showing gains of \textbf{+25.21} (0.5K) and \textbf{+16.34} (2K).
Such large margins indicate that DARFT not only improves greedy decoding accuracy, but also increases the probability quality of the correct answer under stochastic sampling.
Table~\ref{tab:mv20_all} further reports MV@20 results, which aggregate sampled predictions via majority voting.
DARFT achieves substantial gains over SFT in both few-shot settings, improving OA by \textbf{+7.82} and \textbf{+4.81} under 0.5K and 2K data, respectively.
This demonstrates that DARFT leads to more consistent predictions across samples, reflecting increased decision stability and reduced sensitivity to sampling-induced perturbations.

\begin{table}[!t]
\centering
\caption{Comparisons on the QAG-360K test set under \textbf{MV@20} with multi-sample decoding.}
\renewcommand\arraystretch{1.15}
\setlength{\tabcolsep}{6pt}
\begin{tabular}{lcc}
\toprule
\textbf{Method} & \textbf{AA} & \textbf{OA} \\
\midrule
\textbf{Few-shot Fine-tune (0.5K samples)} \\
\midrule
Qwen2.5-VL-3B-SFT   
& 48.12 & 52.41  \\
\rowcolor{gray!12}
Qwen2.5-VL-3B-DARFT 
& 54.04 \chg{+5.92} & \textbf{60.23} \chg{+7.82}  \\
\midrule
\textbf{Few-shot Fine-tune (2K samples)} \\
\midrule
Qwen2.5-VL-3B-SFT   
& 52.94 & 56.42  \\
\rowcolor{gray!12}
Qwen2.5-VL-3B-DARFT 
& 55.53 \chg{+2.59} & \textbf{61.23} \chg{+4.81} \\
\bottomrule
\end{tabular}
\label{tab:mv20_all}
\end{table}

\section{Conclusion}
This work studies CDVQA from the perspective of \emph{decision ambiguity}.
We show that many failures of SFT-adapted vision--language models arise from ambiguous decision boundaries, where the correct option and strong distractors receive highly similar confidence, especially in fine-grained change reasoning.
To address this issue, we define \emph{Decision-Ambiguous Samples} and propose DARFT, a decision-ambiguity-guided reinforcement fine-tuning framework.
DARFT explicitly optimizes relative preferences among competing answers under the same input, sharpening decision boundaries without additional supervision.
Experiments on QAG-360K demonstrate consistent improvements over SFT baselines, with pronounced gains in low-data regimes and ambiguity-prone categories.

\bibliographystyle{IEEEbib}
\bibliography{icme2026references}

\end{document}